\documentclass{article}

\usepackage{arxiv}

\usepackage[utf8]{inputenc} 
\usepackage[T1]{fontenc}    
\usepackage{hyperref}       
\usepackage{url}            
\usepackage{booktabs}       
\usepackage{amsfonts}       
\usepackage{nicefrac}       
\usepackage{microtype}      

\usepackage{pdfpages}
\usepackage{subfigure}
\usepackage{threeparttable}
\usepackage{graphicx} 
\usepackage{epstopdf}
\usepackage{amsmath}
\usepackage{amssymb}
\usepackage{authblk}
\usepackage{natbib}
\usepackage{color}
\definecolor{TRED}{RGB}{195,39,43} 
\definecolor{TGREEN}{RGB}{5,119,72} 
\definecolor{TYELLOW}{RGB}{205,205,0} 

\title{Dynamic Runtime Feature Map Pruning}

\author[1,4]{Tailin Liang}
\author[1,2]{Lei Wang}
\author[1,2]{Shaobo Shi}
\author[1,2,3]{John Glossner}
\affil[1]{University of Science and Technology Beijing, Beijing 100083, China}
\affil[2]{Hua Xia General Processor Technologies, Beijing 10080, China}
\affil[3]{General Processor Technologies, Tarrytown, NY 10591}
\affil[4]{Corresponding author: tailin.liang@xs.ustb.edu.cn}

\begin{document}
\maketitle

\begin{abstract}
High bandwidth requirements are an obstacle for accelerating the training and inference of deep neural networks. Most previous research focuses on reducing the size of kernel maps for inference. We analyze parameter sparsity of six popular convolutional neural networks - AlexNet, MobileNet, ResNet-50, SqueezeNet, TinyNet, and VGG16. Of the networks considered, those using ReLU (AlexNet, SqueezeNet, VGG16) contain a high percentage of 0-valued parameters and can be statically pruned. Networks with Non-ReLU activation functions in some cases may not contain any 0-valued parameters (ResNet-50, TinyNet). We also investigate runtime feature map usage and find that input feature maps comprise the majority of bandwidth requirements when depth-wise convolution and point-wise convolutions used. We introduce dynamic runtime pruning of feature maps and show that 10\% of dynamic feature map execution can be removed without loss of accuracy. We then extend dynamic pruning to allow for values within an $epsilon$ of zero and show a further 5\% reduction of feature map loading with a 1\% loss of accuracy in top-1. 
\end{abstract}
\section{Introduction}\label{sec:intro}
Deep Neural Networks (DNN) \cite{Lecun2015} have been developed to identify relationships in high-dimensional data. Recent neural network designs have shown superior performance over traditional methods in many domains including handwriting recognition, voice synthesis, object classification, and object detection. Using neural networks consists of two steps - training and inference. Training involves taking input data, comparing it against a ground truth of labels, and then updating the weights of the neurons to reduce the error between the the network's output and the ground truth. Training is very compute intensive and typically performed in data centers, on dedicated Graphics Processing Units (GPUs), or on specialized accelerators such as Tensor Processing Units (TPUs).
\par 
\begin{figure}[thb]
    \centering
    \includegraphics[page=4,width=1\textwidth]{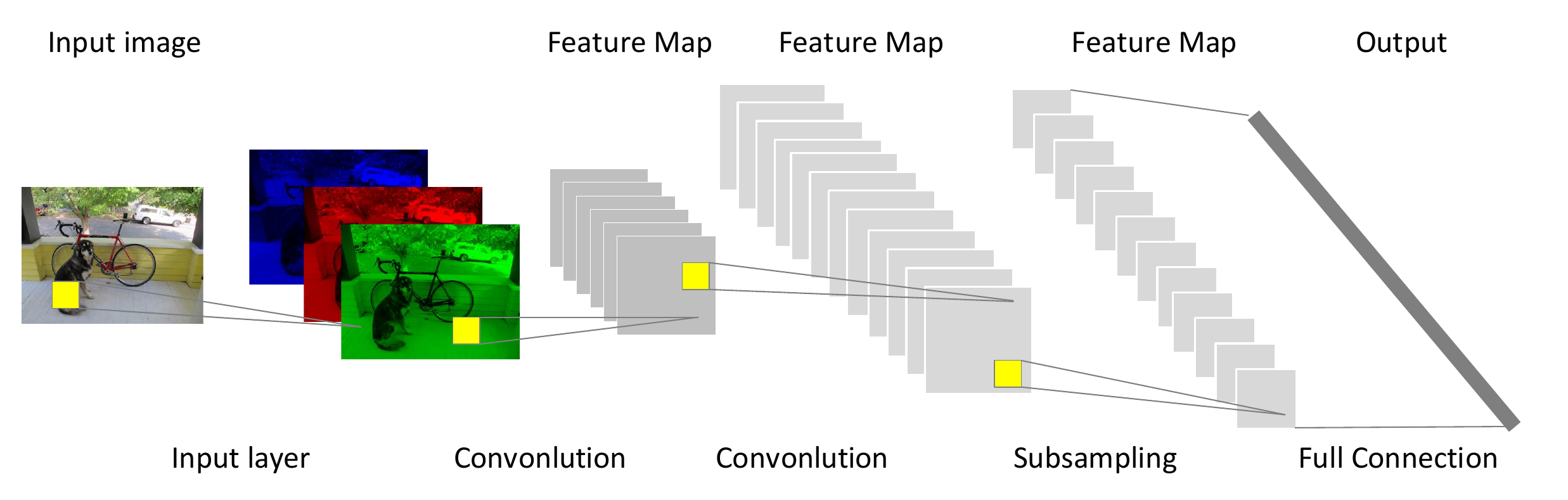}
    \caption{Convolutional Neural Network: Convolutional filters are arranged in layers to detect features in an object. These features "map" the input image to specific characteristics to be predicted from the input data \cite{Lecun2015}.}
    \label{fig:cnn}
\end{figure}
Figure \ref{fig:cnn} shows an image of a dog and a Convolutional Neural Network (CNN) trained to recognize it. A CNN is a class of DNN where a convolution operation is applied to input data. A convolution calculation along with a well trained 3D tensor filter (also known as kernel) can be used identify objects in images. The filters work by extracting multiple smaller bit maps known as feature maps since they "map" portions of the image to different filters. Typically the input pixel image is encoded with separate Red/Green/Blue (RGB) pixels. These are operated on independently and the resulting matrix-matrix multiply for each slice of the tensor is often called a channel \cite{Lecun2015}.
\par
Inference involves processing data on a neural network that has previously been trained. No error computation or back-propagation is traditionally performed in inference. Therefore, the compute requirements are significantly reduced compared to training. However, modern deep neural networks have become quite large with hundreds of hidden layers and upwards of a billion parameters (coefficients) \cite{Iandola2016a}. With increasing size, it is no longer possible to maintain data and parameters in processors caches. Therefore data must be stored in external memory causing significant loading requirements (bandwidth usage). Reducing DNN bandwidth usage has been studied by many researchers and methods of compressing networks have been investigated. Results have shown the number of parameters can be significantly reduced without loss of accuracy. Previous work includes parameter quantization \cite{Micikevicius2017}, low-rank decomposition \cite{Denil2013}, and network pruning which we describe more fully below.
\par
\begin{figure}[htb]
    \centering
    \includegraphics[page=1,width=0.7\textwidth]{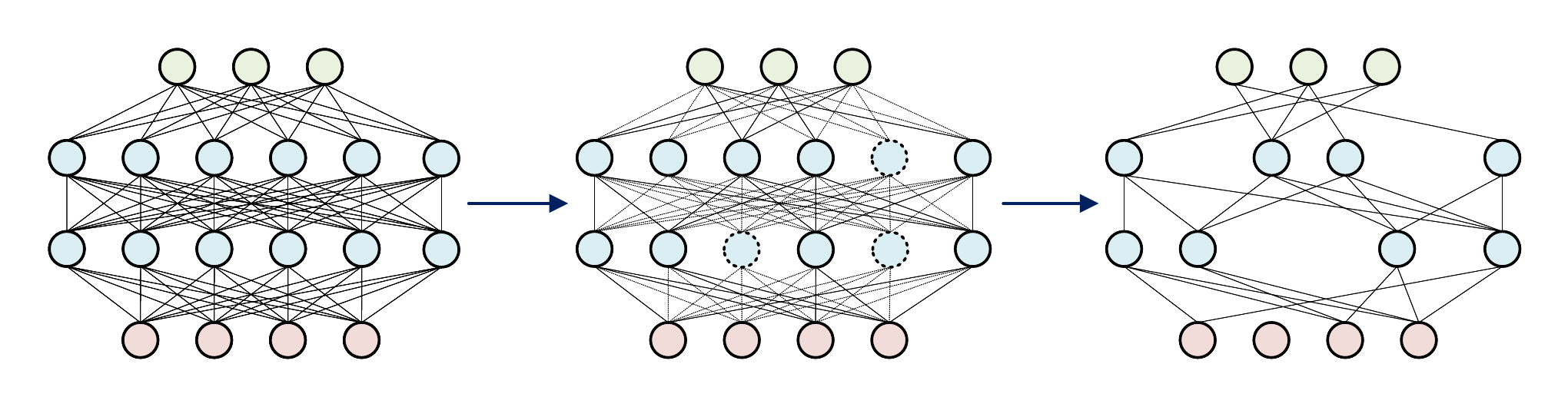}
    \caption{Network Pruning: Neurons are removed judiciously to reduce computations, bandwidth requirements, and power dissipation without significantly affecting the accuracy of the network \cite{Guo2016}.}
    \label{fig:net_pruning}
\end{figure}
Figure \ref{fig:net_pruning} shows an simple example of network pruning. Network pruning involves taking a designed neural network and removing neurons with the benefit of reducing computational complexity, power dissipation, and memory loading. Surprisingly, neurons can often be removed without significant loss of accuracy. Network pruning generally falls into the categories of static pruning and dynamic pruning. 
\par
Static pruning chooses which neurons to remove before the network is deployed. It considers parameter values at or near 0 and removes neurons that wouldn't contribute to classifications. Statically pruned networks may optionally be retrained \cite{Han2015DeepCompression}. While retraining is time consuming it may lead to better performance than leaving the weights as calculated from the unpruned network \cite{Yu2017}. 
With static pruning the pruned models are fixed to an often irregular network structure. A fixed network is also unable to take advantage of 0-valued input data.
\par
Dynamic pruning determines at runtime which neurons will not participate in the classification activity. Dynamic runtime pruning can overcome limitations of static pruning as well as take advantage of changing input data while still reducing parameter loading (bandwidth) and power dissipation. One possible implementation of dynamic runtime pruning considers any parameters that are trained as 0-values are implemented within a processing element (PE) in such a way that the PE is inhibited from participating in the computation \cite{Lin2017}. Sparse matrices fall into this category \cite{Foroosh2015}. 
\par
A kernel map comes from pre-trained coefficient matrices stored in external memory. These are usually saved as a weights file. The kernel is a filter that has the ability to identify input data features. Most dynamic runtime pruning approaches remove kernels of computation \cite{Han2015DeepCompression,han2015,Lecun1989}. In this approach, loading bandwidth is reduced by suppressing the loading of weights. 
\par
Another approach for convolutional neural networks is to dynamically remove feature maps (sometimes called filter "channels"). In this approach channels that are not participating in the classification determination are removed at runtime. This type of dynamic runtime pruning is the focus of this paper. 
\par
In this paper we introduce a method of dynamic runtime network pruning for CNNs that removes feature maps that are not participating in the classification of an object. For networks not amenable to static pruning this can reduce the number of feature maps loaded without loss of accuracy. Retraining is not required and the network preserves Top-1 accuracy. We provide implementation results showing on average a 10+\% feature map loading reduction. We further extend the technique to allow for pruning of feature maps within an epsilon of 0 thereby including networks that use non-zero activation functions. 
\par
This paper is organized as follows. In Section \ref{sec:methodology} we discuss our research methodology. In Section \ref{sec:results} we analyze experimental results. In Section \ref{sec:related} we compare our method with related techniques. In Section \ref{sec:discuss} we discuss the effectiveness of our technique for certain classes of networks and describe our future research. Finally, in Section \ref{sec:conclusion} we conclude and summarize our results. 
%
\section{Runtime Feature Map Pruning}\label{sec:methodology}
%
\begin{figure}[h]
    \centering
    \includegraphics[page=9,width=1\textwidth]{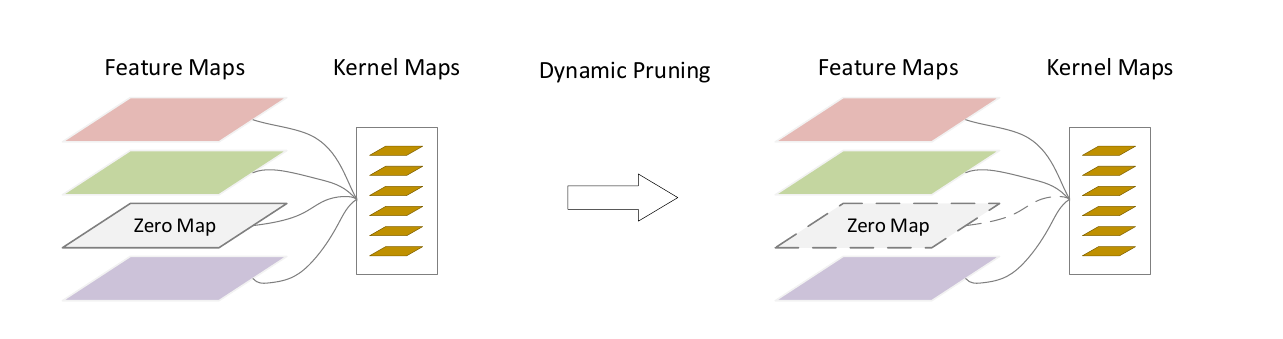}
    \caption{Dynamic Runtime Feature Map Pruning: Prune less contributed feature maps to reduce both data loading and computation resources usage by stopping minor feature maps and corresponding kernel maps goes into computation units}
    \label{fig:dynamic_pruning}
\end{figure}
%
Even with memory performance improvements, bandwidth is still a limiting factor in many neural network designs \cite{Rhu2018}. When direct connection to memory is not possible, common bus protocols such as PCIe further limit the peak available bandwidth within a system. Once a system is fixed, further performance improvements may only be achieved by reducing the bandwidth requirements of the networks being implemented. One way of achieving this is by reducing the number of feature maps being loaded. 
\par
\begin{figure}[htb]
    \centering
    \includegraphics[page=1,width=1\textwidth]{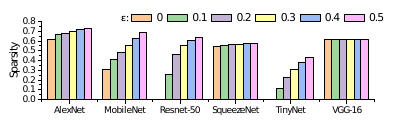}
    \caption{Feature Map Value Sparsity in Convolutional Layers: The x-axis identifies a number of popular CNNs with each bar being the value of coefficients at or within an epsilon of 0. The y-axis gives the proportion of coefficients within that epsilon.}
    \label{fig:sparsity}
\end{figure}
Figure \ref{fig:sparsity} shows the feature map sparsity of some common neural networks. Only convolutional layers are considered (i.e. fully connected layers are not included). The baseline measure of sparsity is 0-valued feature map elements as represented by the orange bar. As an example, AlexNet has over 60\% of feature map with 0 values while ResNet and TinyNet don't have any 0-valued feature map element. It also shows that for values within an epsilon of 0, some networks have many more elements that can be pruned (e.g. MobileNet and TinyNet) while other networks have much smaller variance (e.g. VGG16). 
\par
We also did statistics for all parameters (every single value in trained model) and convolutional coefficients (kernel map) in table \ref{tab:para_spar} and table \ref{tab:kmap_spar}. Basically they are similar except VGG and AlexNet who has heavy number of parameter for connected coefficients. In our experiments, we prune the minor valued kernel map coefficients (both in convolutional layer and connected layer, biases not included) without any fine tune, then valid with 100 images of ImageNet dataset, the result is showing with three colors in table \ref{tab:para_spar}, Green means both Top-1 and Top-5 drop less than 1\%, Yellow indicates one of them (mostly Top-1) drop more than 1\%, Red means both of Top-1 and Top-5 drop more than 1\%.
\par
\begin{table}[]
\caption{All Parameter Sparsity}
\label{tab:para_spar}
\resizebox{\textwidth}{!}{
\begin{tabular}{@{}lrrrrrrrrr@{}}
\toprule
Parameter Sparsity         & 0.000   & 0.005   & 0.010   & 0.020   & 0.040   & 0.060   & 0.080   & 0.100   & 0.200   \\ \midrule
AlexNet                    & {\color{TGREEN}0.00\%}  & {\color{TGREEN}71.29\%} & {\color{TRED}92.15\%} & 98.34\% & 99.74\% & 99.91\% & 99.95\% & 99.97\% & 99.99\% \\
MobileNet                  & {\color{TGREEN}0.00\%}  & {\color{TGREEN}9.02\%}  & {\color{TGREEN}12.56\%} & {\color{TGREEN}19.24\%} & {\color{TGREEN}31.99\%} & {\color{TYELLOW}43.84\%} & {\color{TRED}54.47\%} & 63.76\% & 90.12\% \\
ResNet50                   & {\color{TGREEN}0.00\%}  & {\color{TYELLOW}61.86\%} & {\color{TRED}86.83\%} & 97.29\% & 99.39\% & 99.64\% & 99.72\% & 99.74\% & 99.78\% \\
SqueezeNet                 & {\color{TGREEN}0.00\%}  & {\color{TYELLOW}9.18\%}  & {\color{TYELLOW}18.17\%} & {\color{TYELLOW}35.14\%} & {\color{TYELLOW}62.84\%} & {\color{TRED}80.38\%} & 89.82\% & 94.52\% & 99.35\% \\
TinyNet                    & {\color{TGREEN}0.00\%}  & {\color{TYELLOW}19.71\%} & {\color{TYELLOW}37.70\%} & {\color{TRED}64.25\%} & 86.39\% & 93.05\% & 95.91\% & 97.36\% & 98.95\% \\
VGG16                      & {\color{TGREEN}0.00\%}  & {\color{TYELLOW}87.29\%} & {\color{TRED}96.68\%} & 99.54\% & 99.94\% & 99.97\% & 99.99\% & 99.99\% & 99.99\% \\
\bottomrule
\end{tabular}}
\end{table}
\par
\begin{table}[tbhp]
\caption{Convolution Kernel Sparsity}
\label{tab:kmap_spar}
\resizebox{\textwidth}{!}{
\begin{tabular}{@{}lrrrrrrrrr@{}}
\toprule
Kernel Sparsity       & 0.000   & 0.005   & 0.010   & 0.020   & 0.040   & 0.060   & 0.080   & 0.100   & 0.200   \\ \midrule
AlexNet               & 0.00\% & 29.97\% & 52.66\% & 80.34\% & 95.98\% & 98.60\% & 99.32\% & 99.61\% & 99.94\%  \\
MobileNet             & 0.00\% & 8.85\%  & 12.39\% & 19.11\% & 32.00\% & 44.01\% & 54.80\% & 64.23\% & 90.95\%  \\
ResNet50              & 0.00\% & 62.03\% & 87.09\% & 97.57\% & 99.65\% & 99.89\% & 99.96\% & 99.98\% & 100.0\% \\
SqueezeNet            & 0.00\% & 9.15\%  & 18.12\% & 35.07\% & 62.79\% & 80.36\% & 89.81\% & 94.52\% & 99.35\%  \\
TinyNet               & 0.00\% & 19.90\% & 38.05\% & 64.85\% & 87.19\% & 93.89\% & 96.78\% & 98.22\% & 99.80\%  \\
VGG16                 & 0.00\% & 47.48\% & 77.77\% & 96.17\% & 99.49\% & 99.83\% & 99.93\% & 99.96\% & 99.99\%  \\
\bottomrule
\end{tabular}}
\end{table}
\par
Most current techniques only prune kernel maps. Our technique proposes not to remove entire kernels but only specific feature maps that do not contribute to the effectiveness of the network. This is done dynamically at runtime and has the advantage of reducing the number of feature maps loaded (i.e. bandwidth) without limiting the type of network architecture that can be pruned. 
\par
\begin{figure}[htb]
    \centering
    \includegraphics[width=1\textwidth]{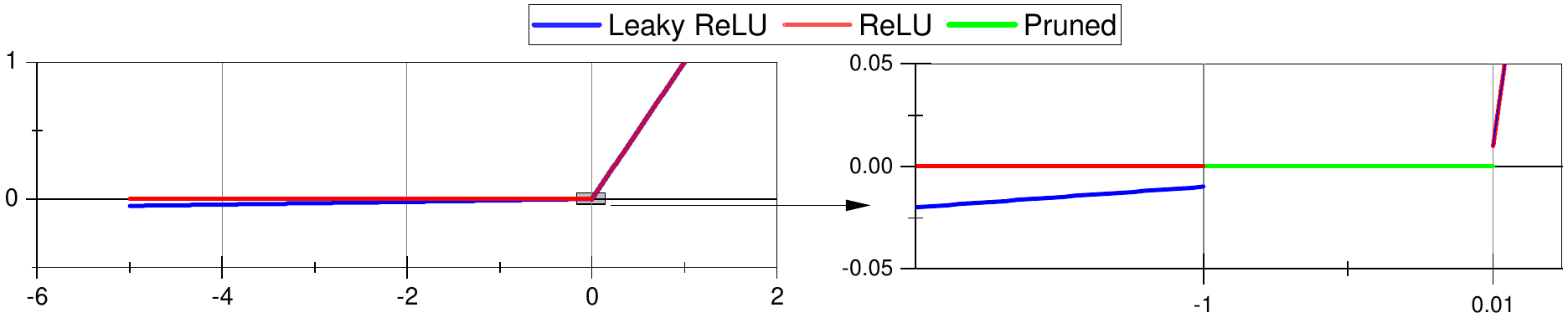}
    \caption{Left: ReLU and Leaky ReLU activation function layout, Right: prune with $|\epsilon|=0.01$ layout}
    \label{fig:actvi}
\end{figure}
\par
Figure \ref{fig:actvi} shows activation functions for ReLU and leaky ReLU. ReLU has the property that for all values of $x \leq 0$ the function remains 0 (e.g. $f(x \leq 0)=0$). Leaky ReLU and some other similar activation functions do not have this property and allow small negative values of $x$ so as to smooth gradients during training \cite{Maas2013}. A result of this is that they have many less 0-valued parameters. ResNet and TinyNet both use leaky ReLU and as shown in Figure \ref{fig:sparsity} they both have low 0-sparsity. However, in some cases, while not exactly zero, the values may be close to 0. 
\begin{equation} \label{eqn:epsilon}
active(x)=\begin{cases}
x & \mbox{ $\epsilon < x $} \\
0 & \mbox{-0.01 $\epsilon \leq x \leq \epsilon$} \\
0.01\times x & \mbox{$x < -0.01 \epsilon$}
\end{cases}
\end{equation}
Equation \ref{eqn:epsilon} shows the activation function we compute for feature maps with epsilon pruning. For any value of epsilon greater than $x$ the function returns $x$. For positive values of $x$ that are less than epsilon the function returns 0 and effectively is pruned. To accommodate leaky ReLU we multiply negative values of $x$ by a small coefficient to transform it to a positive value that is likely to be pruned. 
\par
%
\vspace{1ex}
\begin{tabular*}{\textwidth}{l}
\toprule
    \textbf{Algorithm 1:} Dynamic Feature Map Pruning   \\
\textbf{Input:} channel size, including height, width, number (H, W, C). \\
capability of processor, max width and height to process (h, w). \\
\textbf{Output:} marker for small data filled channels. \\
we define the column ``part + 1'' as the zero mark of each channel. \\
\textbf{for each} i \textbf{in} C \\
\hspace{2ex} // get the channel pieces number\\
\hspace{2ex}channel\_part = ceil ( W $\times$ H / w $\times$ h) \\
\hspace{2ex} // check if all zero in one part\\
\hspace{2ex}\textbf{for all} j \textbf{in} channel\_part \hspace{4ex} \\
\hspace{4ex}   \textbf{ for all} k \textbf{in }w $\times$ h \\
\hspace{6ex}       \textbf{ if }( abs(value[k] < $\epsilon$ ) \\
\hspace{8ex}           channel\_zero\_mark[i][j]=1 \\
\hspace{6ex}        \textbf{end if} \\
\hspace{4ex}    \textbf{end for} \\
\hspace{2ex} // check if all zero in one channel\\
\hspace{2ex}channel\_zero\_mark[i][channel\_part + 1] = sum (channel\_zero\_mark[i][0 : channel\_part]) \\
\hspace{2ex}\textbf{end for} \\
\textbf{end for} \\
\bottomrule
\end{tabular*}
\label{algo}
\par

Algorithm 1 describes a brute-force naive technique for dynamic feature map pruning. It is applied after activation function of Equation \ref{eqn:epsilon} is applied. For all feature map channels in a convolutional neural network we look at the element values. If we determine that a feature map has 0-valued coefficients such that the entire channel is unused, we mark it and subsequently do not compute any values for that feature map. Specifically, we count the number of zeros (or absolute value less than $\epsilon$). If the entire channel is filled with values less than $\epsilon$ we then regard this channel as a zero channel and mark it for later identification. When loading a feature map for processing, if the channel was marked to be within an $\epsilon$ of 0, we will prune it. Some implementations may not implement sufficient neurons (multiply-accumulate units) to process an entire feature map simultaneously. In this case we will break the feature map into smaller pieces. We then sum the flag of each part to determine if the entire feature map is filled with $\epsilon$ zero elements. If so, the entire feature map will be marked as zero-filled and will be skipped thus saving feature map loading and reducing bandwidth requirements. Our source code for our technique is available at Github\footnote{\url{https://github.com/liangtailin/darknet-modified}}.
\section{Experimental Results}\label{sec:results}
\subsection{Classification Accuracy}
We implemented our dynamic pruning algorithm using Darknet - a C language deep neural network framework \cite{darknet13}. We compute statistics by counting the number of feature maps loaded, noting that if a computation unit has a small cache, a feature map will be loaded more than once. In this work we don't consider this additional effect. 
\begin{table}[htbp]
  \centering
  \caption{Accuracy performance of different $\epsilon$ pruning}
  \label{tab:rst_accuracy_loss}
  \resizebox{\textwidth}{!}{
    \begin{tabular}{lrrrrrr|rrrrrr}
    \toprule
          & \multicolumn{6}{c}{Top-1 Accuracy}                     & \multicolumn{6}{c}{Top-5 Accuracy} \\
    \midrule
    Epsilon & \multicolumn{1}{c}{0.0}   & \multicolumn{1}{c}{0.1}   & \multicolumn{1}{c}{0.2}   & \multicolumn{1}{c}{0.3}   & \multicolumn{1}{c}{0.4}   & \multicolumn{1}{c}{0.5}   & \multicolumn{1}{c}{0.0}   & \multicolumn{1}{c}{0.1}   & \multicolumn{1}{c}{0.2}   & \multicolumn{1}{c}{0.3}   & \multicolumn{1}{c}{0.4}   & \multicolumn{1}{c}{0.5}  \\
    \midrule
    AlexNet & 57.17\% & -0.01\% & 0.01\% & 0.07\% & 0.12\% & 0.14\% & 80.20\% & -0.01\% & 0.00\% & 0.02\% & 0.03\% & 0.08\% \\
    MobileNet & 71.44\% & -0.02\% & 0.02\% & 0.74\% & 3.97\% & 24.64\% & 90.35\% & 0.01\% & 0.06\% & 0.49\% & 2.46\% & 19.91\% \\
    ResNet-50 & 75.83\% & 0.02\% & 0.06\% & 1.13\% & 5.56\% & 12.22\% & 92.89\% & 0.00\% & 0.03\% & 0.68\% & 3.26\% & 7.81\% \\
    SqueezeNet & 57.13\% & 0.00\% & 0.00\% & 0.00\% & 0.00\% & 0.00\% & 80.13\% & 0.00\% & 0.00\% & 0.00\% & 0.00\% & 0.00\% \\
    TinyNet & 58.71\% & 0.00\% & 0.01\% & 0.07\% & 1.28\% & 6.96\% & 81.73\% & 0.00\% & 0.00\% & 0.04\% & 0.90\% & 5.99\% \\
    VGG-16 & 70.39\% & 0.00\% & 0.00\% & 0.00\% & 0.00\% & 0.00\% & 89.79\% & 0.00\% & 0.00\% & 0.00\% & 0.00\% & 0.00\% \\
    \bottomrule
    \end{tabular}}%
\end{table}%

To validate our technique we used the ILSVRC2012-50K image dataset containing 1000 classes \cite{Russakovsky2015}. Table \ref{tab:rst_accuracy_loss} lists our results based on the 50,000 images using GPU acceleration.
\par
Our results show that using an $\epsilon = 0.1$ for pruning has no significant loss in accuracy for the top-1 while reducing feature map loading up to 10\%. At $\epsilon=0.2$, top-1 accuracy drooped but top-5 accuracy is still state-of-the-art performance. We note that not all networks improved using this approach. We comment on that in Section \ref{sec:discuss}.
The results show that layers using ReLU activation have the most feature maps removed and therefore the highest feature map loading reduction. Leaky ReLU with an $\epsilon = 0.0$ has little advantage as its feature maps do not have many 0-values.
\par
\begin{table}[htb]
\caption{Accuracy of Networks And Number of Feature Maps Pruned}
\label{tab:rst_per_img}
\centering
\resizebox{\textwidth}{!}{
\begin{tabular}{llllll|lllll|ll}
\toprule
            & \multicolumn{5}{c}{without pruning}             & \multicolumn{5}{c}{$\epsilon = 0.1$ pruning}        & \multicolumn{2}{c}{saved fmap load}    \\ \midrule
net         & cat     & dog     & eagle   & giraffe & horse   & cat     & dog     & eagle   & giraffe & horse   & \multicolumn{2}{l}{saved (avg saved/no prune)}  \\ \midrule
AlexNet     & 40.35\% & 19.03\% & 79.03\% & 36.50\% & 53.13\% & 40.34\% & 19.12\% & 79.15\% & 36.47\% & 53.23\% & 5.51\%  & (43k/781k)   \\
MobileNet   & 24.64\% & 28.51\% & 91.43\% & 27.40\% & 29.35\% & 28.56\% & 29.59\% & 92.02\% & 28.00\% & 28.03\% & 10.16\% & (940k/9255k)   \\
ResNet-50   & 23.77\% & 95.19\% & 68.11\% & 76.37\% & 20.94\% & 19.84\% & 94.89\% & 68.06\% & 73.38\% & 20.63\% & 1.60\%  & (323k/20086k)   \\
SqueezeNet  & 92.67\% & 57.34\% & 58.60\% & 45.58\% & 93.17\% & 92.67\% & 57.34\% & 58.60\% & 45.58\% & 93.17\% & 0.56\%  & (18k/3290k)   \\
TinyNet     & 15.25\% & 14.51\% & 54.11\% & 29.71\% & 25.59\% & 15.25\% & 14.51\% & 54.11\% & 29.71\% & 25.59\% & 0       & (0/3458k)   \\
VGG-16      & 26.79\% & 56.49\% & 92.12\% & 97.73\% & 39.04\% & 26.82\% & 56.55\% & 92.14\% & 97.73\% & 39.01\% & 0.32\%  & (50k/15087k)   \\
 \bottomrule
\end{tabular}}
\end{table}
Table \ref{tab:rst_per_img} shows single image accuracy without pruning and pruning with $\epsilon=0.1$. We note that the labels are not changed (i.e. there is no prediction involved). We compare the ground truth labels with the effects exclusively related to pruning the network. In some cases the pruned network outperformed the unpruned. This is inline with other researcher's results \cite{Yu2017,Huang2018a}. The last column shows the number of feature maps pruned at $\epsilon=0.1$. For example, MobileNet reduced the number of feature maps loaded by 940 thousand out of a total number of 9255 thousand feature maps to be computed. This is approximated a 10\% savings in the number of feature maps loaded. It should be noted that due to long simulation times Table \ref{tab:rst_per_img} was determined using 5 random images. Therefore the results should be considered preliminary.
\par
\par
\begin{figure}[htb]
    \centering
    \includegraphics[width=0.85\textwidth]{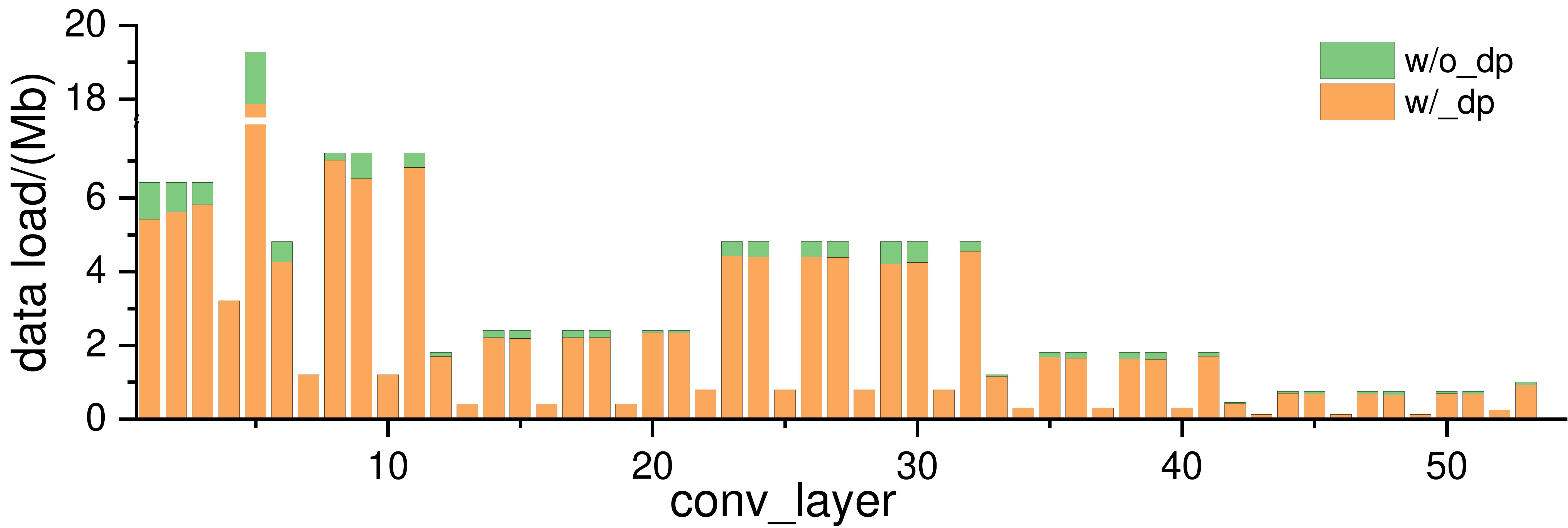}
    \caption{Dynamic Pruning Performance on MobileNet by Convolutional Layer}
    \label{fig:bw-mobile}
\end{figure}

Additional results for MobileNet not shown in Table \ref{tab:rst_per_img} reveals that MobileNet particularly benefited from feature map pruning. With $\epsilon = 0$ pruning, MobileNet reduced 36/54 ReLU activated convolutional layers resulting in a feature map loading reduction of 7.8\%. AlexNet reduced 3/5 ReLu activated convolutional layers reduced feature map loading of 5.1\%, while SqueezeNet reduced 5/26 layers resulting in a 0.7\% reduction. Other networks using leaky ReLU, as anticipated, do not have reduced feature map loading with $\epsilon = 0$. Figure \ref{fig:bw-mobile} shows the convolutional layer-by-layer feature map loading with and without dynamic pruning. 
\par
Among the networks we've tested, MobileNet and AlexNet (which both use ReLU) are much improved while Squeezenet (uses ReLU), VGG-16 (uses ReLU) and Resnet-50 (leaky and linear) are marginally if at all improved. Finally, Tinynet (leaky ReLU) is not improved at all.
\par
\par
Figure \ref{fig:bw-mobile} shows the feature loading requirements for MobileNet by convolution layer. The y-axis displays the Mega-bits of data required to be read. The x-axis displays the network layers. The stacked bars show the data requirements with and without dynamic pruning. $\epsilon=0$ is used and shows that dynamic pruning can reduce the image "dog", which shown in \ref{fig:cnn}, data loading requirements by about 9.2\% as averaged across all the layers. A few layers of MobileNet use linear activation functions and therefore don't benefit from $\epsilon=0$ pruning.
\par
We've also run AlexNet experiments on Caffe \footnote{\url{https://github.com/BVLC/caffe}} with \footnote{\url{https://github.com/songhan/Deep-Compression-AlexNet}} and without \footnote{\url{http://dl.caffe.berkeleyvision.org/bvlc_alexnet.caffemodel}} static pruning. The results shows a similar runtime feature map sparsity of about 60\%. Resulting 0.45\% feature map loading reduction with channel-wise dynamic feature map pruning after static pruning is applied, the result for none static pruned is also minor as 0.79\%.
\begin{table}[htb]
\caption{Top-1 Accuracy Loss Versus $epsilon$ Pruning Threshold} 
\label{tab:rst_epsilon_step}
\resizebox{\textwidth}{!}{\begin{tabular}{lll|ll|ll|ll|ll|ll}\toprule
$\epsilon$ & \multicolumn{2}{c}{0.0} & \multicolumn{2}{c}{0.1} & \multicolumn{2}{c}{0.2} & \multicolumn{2}{c}{0.3} & \multicolumn{2}{c}{0.4} & \multicolumn{2}{c}{0.5} \\ \midrule
net        & accuracy    & reduce    & loss       & reduce     & loss      & reduce      & loss        & reduce    & loss      & reduce      & loss       & reduce      \\ \midrule
AlexNet    & 57.173\%    & 5.064\%   & 0.050\%    & 5.885\%    & 0.050\%    & 6.552\%    & 0.100\%    & 7.951\%    & 0.450\%    & 9.174\%    & 0.450\%    & 10.066\%   \\
MobileNet  & 71.449\%    & 7.796\%   & -0.250\%   & 10.297\%   & -0.350\%   & 10.863\%   & 0.450\%    & 11.338\%   & 4.450\%    & 12.268\%   & 29.250\%   & 14.386\%   \\
SqueezeNet & 57.129\%    & 0.707\%   & 0.000\%    & 0.703\%    & 0.000\%    & 0.714\%    & 0.000\%   & 0.716\%    & 0.000\%    & 0.717\%    & 0.000\%   & 0.718\%    \\
\bottomrule

\end{tabular}}
\end{table}
\subsection{Balance Between Feature Map Loading Reduction and Accuracy}
We characterized a range of $\epsilon$ values between 0 and 0.5 with a step size of 0.1 to determine if a feature map should be pruned. We evaluated the effect of $\epsilon$ on MobileNet, SqueezeNet, and TinyNet using 2000 images from ImageNet. The results show that as $\epsilon$ increases the accuracy decreases. Table \ref{tab:rst_epsilon_step} shows the accuracy loss with varying $\epsilon$. %
\section{Related Work}\label{sec:related} 
Bandwidth reduction is accomplished by reducing the loading of parameters or the number of bits used by the parameters. This lead to further research on sparsity, quantification, and weight sharing.
Bandwidth reduction using reduced precision values has been previously proposed. Historically most networks are trained using 32-bit single precision floating point numbers \cite{Sze2017a}. Ma \cite{Ma2016} and Kaiming \cite{Kaiming2015} have shown that 32-bit single precision parameters produced during training can be reduced to 8-bit integers for inference without significant loss of accuracy. Jacob \cite{Jacob2017} used 8-bit integers for both training and inference but with an accuracy loss of 1.5\% on ResNet-50. Li \cite{Li} and Leng \cite{Leng2017} showed that for ternary weight-only quantized networks the accuracy drop was as low as 2.57\%. Zhou \cite{Zhoua2016}, and Hubara \cite{Hubara2016a} took this to an extreme using only 1-bit binary weights, less than 4-bit activation functions, and 6-bit gradients with accuracy losses of 16.4\%. Zhou expanding on his earlier work \cite{Zhou2017} and Das \cite{Das2018} improved the accuracy loss to between 0.49\% to 2.28\% using binary values for pre-trained networks and fine tuning the results. Our present work does not consider reduced precision parameters but may be incorporated into future research since it is complementary to our approach.
\par
Compressing sparse parameter networks has been carried out to save both computation and bandwidth, 
Chen \cite{chen2015} describes using a hash algorithm to decide which weights can be shared. This focused only on fully connected layers and uses pre-trained weight binning rather than dynamically determining the bins during training. Han \cite{Han2015DeepCompression} describes weight sharing combined with Huffman coding. Weight sharing is accomplished by using a k-means clustering algorithm instead of a hash algorithm to identify neurons that may share weights. This realized a pruned and quantized parameter storage reduction of $1.18\times$ to $1.58\times$. In our technique we don't currently share weights but it is possible to combine our technique with weight sharing.\par
Studies by Foroosh and Glorot \cite{Foroosh2015,Glorot2011} have shown that in networks with many parameters a significant amount of the parameters have values of 0 or close to zero. Foroosh feeds the sparse maps into a sparse matrix multiplication algorithm which decomposes the maps into a sparse format to accelerate the computation. This technique shows an accuracy loss of less than 1\%. Ren \cite{Ren2018} exploits this sparsity using low-resolution segmentation networks. They compute convolutions on a block-wise decomposition of the mask. This technique is used in an autonomous vehicle application where lidar scans are shown to have repetition and 0-valued numbers.
\par 
Han \cite{Han2016a} further incorporates weight sharing and sparsity into his Efficient Inference Engine (EIE) \cite{Han2016}. The EIE is intended for fully connected layers where the shared and compressed (sparse pruned) nets are used for efficient inference.
Lavin \cite{Lavin2015} further shows that a $3\times3$ winograd convolution can reduce throughput for small filter sizes by $2.25\times$. EIE could support this kind of convolution by tuning the channel-wise convolution to matrix $\times$ vector operations.
\par
Network pruning is an important component for both memory size and bandwidth usage. It also reduces the number of computations. Early research used large scale networks with static pruning to generate smaller networks to fit end-to-end applications without significant accuracy drop \cite{Bucilua2006}.
\par
LeCun as far back as 1990 proposed to prune non-essential weights using the second derivative of the loss function \cite{LeCun}. This static pruning technique reduced network parameters by a quarter. He also showed that the sparsity of DNNs can provide opportunities to accelerate network performance. 
\par
Han \cite{Han2015DeepCompression} described a static method to prune kernel maps that don't contribute to the classification. This technique uses re-training to optimize the neural network. They also specifically remove \textbf{weights} with small values in addition to feature map pruning. The pruning and retraining process was iterated 3 times for a $9\times$ reduction in total parameters.
Han's learning weights and connection work requires retraining. Guo \cite{Guo2016} describes a method using pruning and splicing that compressed AlexNet by factor of $7.7\times$. This significantly reduced the training iterations from 4800K to 700K. However this type of pruning results in an asymmetric network complicating hardware implementation. 
\par
Most network pruning methods typically prune the kernel rather than feature maps \cite{Guo2016}. In addition to significant retraining times, most of the weight compression is contributed by fully connected layers. AlexNet and VGG particularly have many parameters in fully connected layers. Convolutional layers by contrast realize only a 50\% reduction on average \cite{Han2015DeepCompression}. Our technique uses dynamic pruning of feature maps rather than weights and requires no retraining. We realize a 10.2\% reduction in feature map loading without loss of accuracy and importantly we do not limit the type of networks that can be pruned.
\par
Bolukbasi \cite{Bolukbasi2017AdaptiveNN} has reported a system that can adaptively choose which layers to exit early. They format the inputs as a directed acyclic graph with various pre-trained network components. They evaluate this graph to determine leaf nodes where the layer can be exited early. This can be considered a type of dynamic layer pruning. 
\par 
For instruction set processors, Feature maps or the number of filters used to identify objects is a large portion of bandwidth usage \cite{Sze2017} - especially for depth-wise or point-wise convolutions where feature map computations are a larger portion of the bandwidth\cite{Chollet2017}. 
Lin \cite{Lin2017} used Runtime Neural Pruning (RNP) to pretrain a side network to predict which feature maps wouldn't be needed. This is a type of dynamic runtime pruning. They found $2.3\times$ to $5.9\times$ acceleration with top-5 accuracy loss from 2.32\% to 4.89\%. 
However, in this way of pruning, the pre-trained supervision network still needs to load parameters to figure out which feature maps will be removed. In our approach an additional neural network to predict which feature maps may be ignored isn't required. Instead, we look for feature maps that are not being used in the current classification. RNP, as a predictor, may need to be retrained for different classification tasks. The prediction network may also increase the original network size. Our technique doesn't require additional neurons (or networks) to determine if a feature map will be useful to the current classification (e.g. it is not a prediction). 

Rather, we look at all the feature maps and remove the maps that are dynamically determined not to be participating in the classification.
\par
%
Rhu \cite{Rhu2018} recently described a compressing DMA engine (cDMA) that improved virtualized DNNs (vDNN) performance by 32\%. It compresses activated feature maps using ReLU activation functions prior to transfer. The current implementation operates on elements and takes advantage of sparsity within a feature map without removing the entire channel. The technique uses a bitmap to record all non-zero/zero elements of a feature map. The information is then compressed for efficient transfer across a PCIe bus. Compressed sparse maps with a zero bitmap are then decompressed using a function similar to Caffee's im2col \cite{Jia2014}. A disadvantage of this approach is that it spreads out the feature map making it difficult to map to computation units.
Our technique prunes by channel rather than elements. This benefits instruction set processors, particularly signal processors, because data can be easily loaded into the processor using sliding windows. 
\section{Discussion and Future Work}\label{sec:discuss}
In our experiment, we count feature map loading once as we assume the system has sufficient memory to hold all the feature map weights and data. For some networks (e.g. VGG-16) that may require 60MB just for the feature map weights. For processors with less capacity this will require the maps to be partially loaded. We plan to investigate this in future works.
\par
From our experiments we conclude that for ReLU activated neural networks (e.g. AlexNet) static pruning can be effective since ReLU converts negative values to zeros leading to sparse networks. We did find minor improvements of <0.5\% when applying dynamic pruning after static pruning with Caffe. However not all networks can be statically pruned (e.g. ResNet and TinyNet).
\par
\begin{figure}[htb]
    \centering
    \begin{subfigure}[Normal Convolution]{
        \includegraphics[width=0.45\textwidth,page=7]{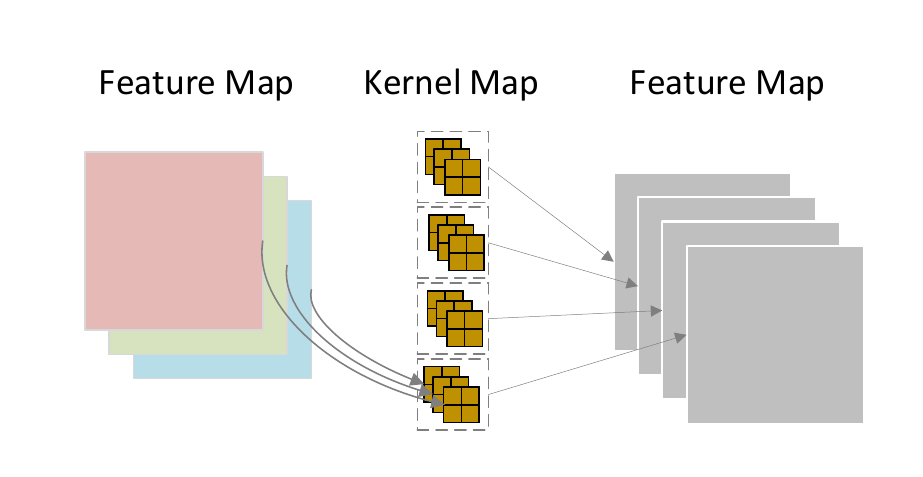}
        \label{fig:diff_conv:normal}}
    \end{subfigure}
    \begin{subfigure}[Depth-wise Convolution]{
        \includegraphics[width=0.45\textwidth,page=8]{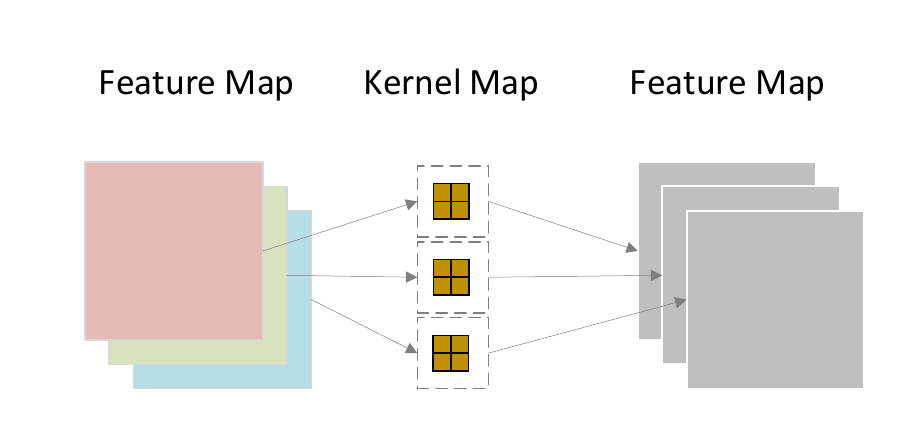}
        \label{fig:diff_conv:depth_wise}}
    \end{subfigure}
    \caption{Comparing Depth-wise Convolution with Normal Convolution}
    \label{fig:diff_conv}
\end{figure}

Figure \ref{fig:diff_conv} shows two CNN network architectures - one using traditional convolutional kernels (figure \ref{fig:diff_conv:normal}) and one using depth-wise convolutional kernels (figure \ref{fig:diff_conv:depth_wise}). Modern CNNs tend to use depth-wise convolutional kernels due to reduced computational complexity \cite{Chollet2017}. A traditional convolutional layer for an $I$ input, $O$ output, $K$ kernel, $A$ area has $I\times K \times O \times A$ computations required. A depth-wise convolutional layer requires only $I \times K \times A$ computations, Depth-wise convolutional layers, in addition to requiring less computations, have the advantage that the feature map used only once to generate new ones thereby reducing feature map loading. Significantly, with the reduced kernel size, they also increase the ratio of feature maps to kernel maps (weight) by a factor of $O$. We expect this to be of benefit in dynamic feature map pruning. Point-wise convolutions have a similar benefit due to the reduction in weights.
\par
\begin{figure}[h]
    \centering
    \includegraphics[page=6,width=0.9\textwidth]{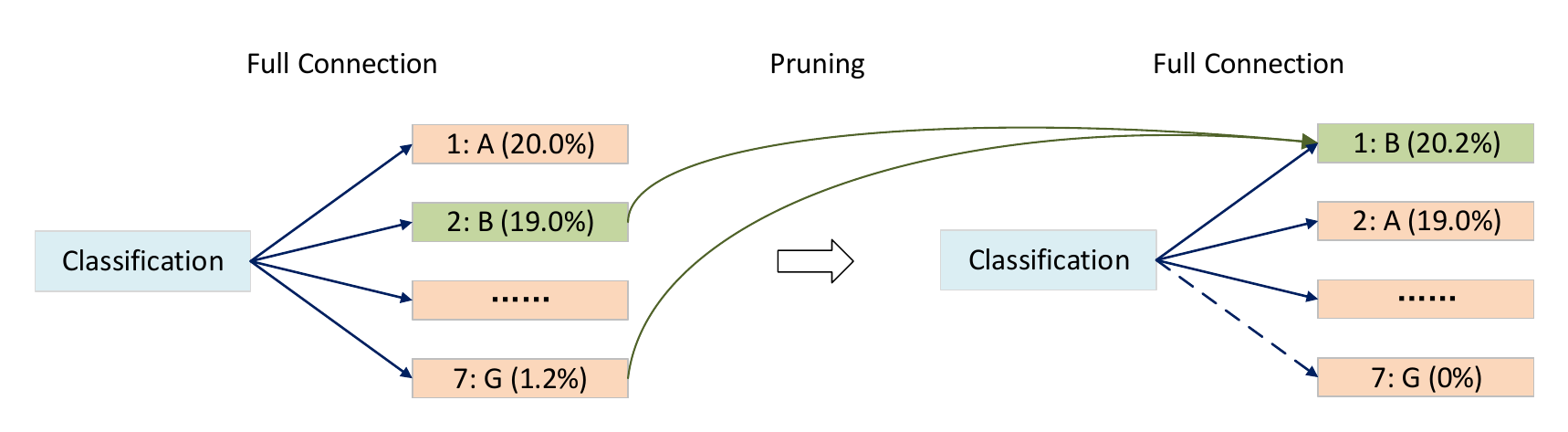}
    \caption{Prune Increased Classification Accuracy: The pruning operation move a small prediction (class G) to a right class (class B) so that the classification result get corrected}
    \label{fig:acc_raise}
\end{figure}

Table \ref{tab:rst_epsilon_step} shows that for some networks an $\epsilon=0.2$ feature map pruning had no loss of accuracy for top-1. As shown in figure \ref{fig:acc_raise}, we believe that is because in the latter part of the network, part of the scattered weights have been transferred to the filter with the highest probability of predicting the correct classification. This is an area of future research.
\par
As we introduced in Section \ref{sec:related}, compression networks typically operate on kernel maps (weights). The more network weights that are statically pruned, the more bandwidth will be consumed by feature maps. Further, unless statically pruning removes an entire neuron, it can not reduce bandwidth usage since the zero feature maps will still be loaded. We have shown that even without weight optimizations, feature map bandwidth is comparable to weight bandwidth when depth-wise (or point-wise) convolutions are employed. Additionally, weight pruning tunes the ratio of weights to feature maps for each layer to balance accuracy and compression. This requires retraining each time a layer is pruned. As networks become deep with many layers, retraining after pruning each layer is computationally expensive. 
\par
As static pruning provides 50\% or less contribution to convolutional layer weight compression, they don't hurt the high runtime sparsity of up to 70\%, with our pruning activation the feature map sparsity going up with minor or without loss of accuracy. The 10\% all-zero feature map also provides additional opportunity of kernel map pruning.
\par
In this work we have determined which feature maps to prune using a fixed $epsilon$. Our future work focuses on determining and possibly dynamically modifying these values by inspecting layer channels. This technique might also be useful during training. 
\begin{figure}[htb]
    \centering
    \includegraphics[page=5,width=0.7\textwidth]{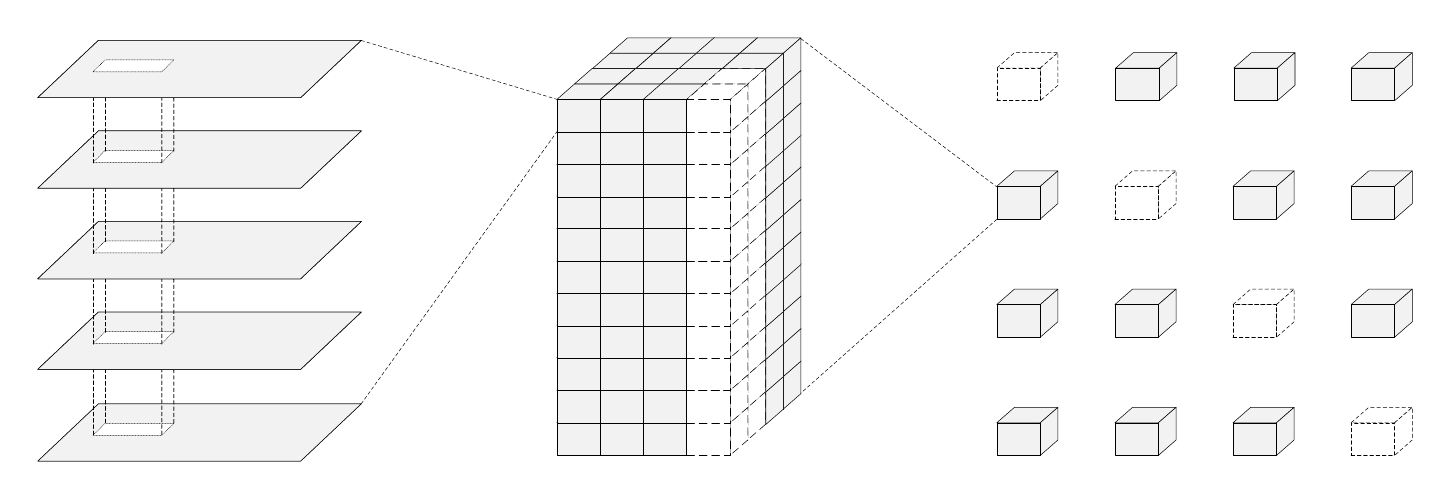}
    \caption{Sparsity distribution of weights: Zeros may be distributed through feature maps (left), tensors (middle), or tensor arrays (right)}
    \label{fig:diff_sparsity}
\end{figure}
\par
As we see in figure \ref{fig:sparsity} some CNNs have 50+\% sparsity. In such cases we found only 10\% channel-wise reduction. We suspect that in a well trained classification model, among the large number of layers and channels, there should be specific maps that predict the final class. Our future work will look specifically at designing networks that take advantage of dynamic feature map pruning where few parameters are 0-valued but feature maps may dynamically be 0-valued. We suspect this may also be of benefit to capsule networks \cite{Sabour2017}. 
%
\section{Conclusion}\label{sec:conclusion}
In this paper, we analyzed feature map parameter sparsity of six different convolutional neural networks - AlexNet, MobileNet, ResNet-50, SqueezeNet, TinyNet, and VGG16. We found a range of sparsity from no sparsity to greater than 50\% sparsity. When considering parameter values an $epsilon$ away from 0, all networks exhibited some level of sparsity. Of the networks considered, those using ReLU (AlexNet, SqueezeNet, VGG16) contain a high percentage of 0-valued parameters (50\%+) and can be statically pruned. However static pruning can lead to irregular networks. Networks with Non-ReLU activation functions in some cases may not contain any 0-valued parameters (ResNet-50, TinyNet). Further static pruning on large networks that require retraining may not be computationally feasible when $epsilon$ values near 0 are considered.
\par
We also investigated runtime feature map usage and found that input feature maps comprise the majority of bandwidth requirements when depth-wise convolution and point-wise convolutions used. 
Our approach uses dynamic runtime pruning of feature maps rather than parameters. This technique is complimentary to static pruning and doesn't require retraining of the CNN. Using this technique we show that 10\% of dynamic feature map execution can be removed without loss of accuracy. We then extend dynamic pruning to allow for values within an $epsilon$ of zero and show a further 5\% reduction of feature map loading with a 1\% loss of accuracy in top-1. We achieved a slight further reduction on networks that were able to be statically pruned. As depth-wise and point-wise convolutional kernels become more common, the amount of computations performed by feature maps will increase possibly further benefiting from dynamic pruning. 

\bibliographystyle{unsrt}
\bibliography{references}
\end{document}